# Automated Switching System for Skin Pixel Segmentation in Varied Lighting

Ankit Chaudhary
ankit-chaudhary@uiowa.edu

Ankur Gupta
f2009087@bits-pilani.ac.in

**ABSRACT-** In Computer Vision, colour-based spatial techniquesoften assume a static skin colour model. However, skin colour perceived by a camera can change when lighting changes. In common real environment multiple light sources impinge on the skin. Moreover, detection techniques may vary when the image under study is taken under different lighting condition than the one that was earlier under consideration. Therefore, for robust skin pixel detection, a dynamic skin colour model that can cope with the changes must be employed. This paper shows that skin pixel detection in a digital colour image can be significantly improved by employing automated colour space switching methods. In the root of the switching technique which is employed in this study, lies the statistical mean of value of the skin pixels in the image which in turn has been derived from the Value, measures as a third component of the HSV. The study is based on experimentations on a set of images where capture time conditions varying from highly illuminated to almost dark.

*Keywords – Skin Detection, Robust Segmentation, Colour Space Switching, Image Processing, Automated Skin Detection*

## 1. INTRODUCTION

Skin Segmentation is gaining importance today due to its wide use in Image processing and its application in virtuallyevery field. Image segmentation is a crucial first step for object recognition, image registration, compression etc. Skin segmentation is commonly used in algorithms for face detection [2][5][10][12], hand gesture analysis [6][14], and objectionable image filtering [16]. In these applications, the search space for objects of interest such as faces or hands can be reduced through the detection of skin regions. To this end, skin segmentation is very effective because it usually involves a small amount of computation and can be done regardless of pose. In colour images, additional colour information available can be utilized. The HSV model closely resembles the human visual systems perception of colour. This model employs the polar co-ordinate space. The Red, Green and Blue (RGB) model is based on physical interpretation of colour. Various other models like $YC_bC_r$ and CMY models are also popular.Most skin segmentation techniques involve the classification of individual image pixels into skin and non-skin categories on the basis of pixel colour. The rationale behind this approach is that the human skin has very consistent colourswhich are distinct from the colours of many other objects. Colour is a powerful cue that can be used as a first step in skin detection because of its advantages: lowcomputational cost, robustness against illumination changing and geometrical transformation.

However skin colour changes with change in illumination and failures are certain if a fixed skin model is used. One of the solutions is colour model adaption [13]. This model was used for tracking skin colourunder varying illumination, viewing geometry and camera parameters. Observed log-likelihood measurements were usedto perform selective adaptation.Object's colour distributions were modelled using Bayesian mixture models in hue-saturation space and an adaptivelearning algorithm was used to update these colour models.Another approach to account for the effects of light source and the high-brightnessregions in colour images, we use the assumptive method of Gray World [17] in order to compensate the images with lightinterference. Based on the theory of dual-colour reflection model [18] and the facts which surface reflection mainly comes from the real colour oflight source, the determination of whether the images have the light interference has been made in [19]. If the image has light interference, light-compensation should be adopted tocompensate for varying illumination colours [20][26]. The whole algorithm is discussed in detail in [5].

## 2. BACKGROUND WORK

In the past few years a number of comparative studies of colour space selection have been reported. Among various types of skin detection methods, the ones that make use of the skin colour as a tool for thedetection of skin is considered to be the most effective [23]. Human skins have a characteristic colour and it was a commonly accepted idea driven by logic to design a method based onskin colour identification. The problem arose with the provision of different varieties of human skinfound in different parts of the world. A number of published researches included various skin models and detection techniques such as [23-24]; however none came up with complete accuracy. Huang [1] used ASSM which used possible combinations of three skin models; ASSM is constructed using the possible combinations of threeskin colour models i.e. $YC_bC_r$ model, Soriano's [9] model and the Bayesian mixture model. Perumal [2] used different probabilistic filter models in six colour spaces and came up with a proposed range of values, that the 'relevant' planes of the model, should lie for detecting skin pixels [3].

## 3. SWITCHING TECHNIQUES

In this section we describe the three techniques that we used to switch between colour spaces. These techniques depend largely on properties of image such as Hue-Value, R-G-B values etc and their functions. The techniques are hereby called Algorithms and the following subsections describe each one of them in detail. In the first technique, first choose the colour space that is suitable to detect skin-like pixels in the given input image(only out of RGB, HSV, and YCbCr). This is done by training an Artificial Neural Network. The detailed algorithm and architecture of the neural network is given later. Then Bayesian routine [7] (3.A) is applied on the most suitable colour space thus obtained.

In the second technique we apply Bayesian routine in the three colour spaces and then calculate the number of pixels in the largest blob of output from each of the colour spaces which is a logical image. The colour space in which largest of these blobs is found is taken as the most suited colour space for detecting skin-like colours. This is done under the assumption that human in the image is closest to the camera while taking the image. Details are given in section 3.C. In the third technique Bayesian routine applied to all the colour spaces to the given input image and then non-weighted addition of all the logical images, which is output from Bayesian routine, is done. Detailed algorithm is given in section 3.D.

### A. Prerequisite to each of the Algorithms(hereby called BayesianRoutine) given below:

The image taken is then converted to the corresponding image map. Now once image map is returned individual pixels are picked and analysed to fit in the test range. Newton's S-R Method is employed to reach to an optimum range for the colour space in consideration. The range(s), for the colour spaces under consideration, so obtained will be referred to as FilterRGB, FilterHSV and FilterYC$_r$C$_b$. The range filter obtained is applied to each pixel in a plane and its corresponding pixels in other two colour planes. Set the pixels which pass the filter to 1 and all other to 0. Thus we get a logical image where white portion roughly maps to the skin portion of the Input image.

Table 1: RESULTS FROM BAYESIANROUTINE

| Colour Space | Conditions |
|---|---|
| RGB | R : 95-255<br>G: 40-255<br>B:20-255 |
| HSV | H:0.04 – 0.0882<br>S:0.11 – 0.68<br>V:0.38 – 0.112 |
| YCbCr | Cb: 100 – 125<br>Cr: 135 - 170 |

For each pixel in the logical image which equals 1 look for the 8 nearest pixels around it and if any of those turn out to be 1, go that pixel and repeat the process and until none of 8 closest pixels is 1.Store the value of number of 1's in the routine. Repeat the above procedure for all pixels. Retrieve the index of the subMatrix mapping to the largest value of connected matrix. Call this subMatrix MaxConnected. Set all the pixels in the logical image equal to 0 except of MaxConnected.The logical matrix is used for allalgorithms.This logical matrix is then mapped back into original given coloured image map. Table 1 shows the results found by the BayesianRoutine.An abstraction for the Bayesian Routine for the three colour spaces in Figure1.

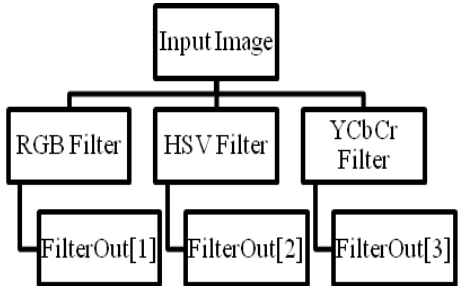

**Fig1** Flow Diagram depicting the abstraction of BayesianRoutine

*B. Switching ColourSpace:Artificial Neural Network (Algorithm1)*

Multilayer networks solve the classification problem for non-linear sets by employing hidden layers, whose neurons are not directly connected to the output. The additional hidden layers can be interpreted geometrically as additional hyper-planes, which enhance the separation capacity of the network. The number of epochs needed to train the network depends on various parameters, especially on the error calculated in the output layer.The output is calculated as follows[22]:

$$y_k(\mathbf{x},\mathbf{w}) = \sigma\left(\sum_{j=0}^{M} w_{kj}^{(2)} h\left(\sum_{i=0}^{D} w_{ji}^{(1)} x_i\right)\right).$$

where the set of all weight and bias parameters have been grouped together into a vector w. Thus the neural network model is simply a nonlinear function from a set of input variables $\{x_i\}$ to a set of output variables $\{y_k\}$ controlled by a vector w of adjustable parameters.

The activation function $\sigma$ used at the output layer is the softmax function for classifying the given image in to one of the three classes taken into account.The softmax function is defined as follows[22]:

$$P_i = \frac{\exp(qi)}{\sum_{j=1}^{n} \exp(qj)}$$

where P is the value of an output node, q is the net input to an output node andn is the number of output nodes. It ensures all of the output values of P are between 0 and 1, and that their sum is 1. The output is assigned the class with maximum probability given the input.

**Precondition for the algorithm:** A large set of image database is collected and filter is applied for all the colour spaces. The best suited colour space isinspected with human supervision which is then stored as the target colour space for the corresponding image. A Neural Network is constructed with input variable as a function of variables

$$f(H_i,S_i,V_i,R_i,G_i,B_i,Y_i,Cb_i,Cr_i),$$
where $i \leq N$, $N = size\ of\ training\ dataset$,

and best suited colour space as the target variable. The undergiven is the pseudoalgorithm for classifying a given image into a suitable colour space for segmenting skin class pixels.

```
1. .A    test    image    is    input    and
   f(Hᵢ,Sᵢ,Vᵢ,,Rᵢ,Gᵢ,Bᵢ,Yᵢ,Cbᵢ,Crᵢ),∀ i,is found.
2. The weights obtained by the neural network
   applied in database described above are
   matrix    multiplied    by    the    matrix
   f(Hᵢ,Sᵢ,Vᵢ,,Rᵢ,Gᵢ,Bᵢ,Yᵢ,Cbᵢ,Crᵢ).
3. The result is the class label i.e. the
   suitable colour space for this image and
   skin segmentation in this colour space is
   applied.
4. The output is hereby called output[0].
```

*C. MaxConnected(Algorithm2):*

This extracts the sub Image after applying the Bayesian Filter in each colourspace and then chooses the Maximum Connected of them all i.e. the one with highest no of pixels in the Region of Interest.The undergiven is the pseudoalgorithm for the same.

```
1.  BayesianRoutine  given  above  is  applied
    for each of the three colour spaces.
2.  MaxConnected[i] is extracted as output
        where i= 0,1,2
        {
                0->RGB,1->HSV,2->YCbCr
        }
3.  For each "i" in {0,1,2}
        i_mPlexed = i multiplexed such that
        MaxConnected[i].NoOfPixels        is
        maximum of all the MaxConnected.
4.  Output=BayesianRoutine  applied  in  the
    colour  space  mapped  by  i,  wherethe
    mapping is given by 1.
5.  The output[1] = Output.
```

*D. SigmaConnect(Algorithm3):*

In this approach, BayesianRoutine given above is applied to each of the three colour spaces and the sum of all the logical sub images is calculated. The logical matrix thus obtained is mapped back onto the original image. The under given is the pseudoalgorithm for the same.

```
1.   BayesianRoutine given above is applied
     for each of the three colour spaces.
2.   for each i in {0,1,2}
     MaxConnected_Master = ∑MaxConnected[i].
3.   MaxColour=MaxConected_Mastermapped  back
     on to the original coloured Image map.
4.   returnoutput[2]=MaxColour.
```

Table 2: HIDDENLAYER-WEIGHT MATRIX FOR THE INPUT TUPLE

| Neuron# / Weight# | 1 | 2 | 3 |
|---|---|---|---|
| 1 | 0.539833 | -0.41466 | 0.128368 |
| 2 | -0.33529 | -0.26512 | 0.292686 |
| 3 | -0.32967 | -0.55298 | 0.229643 |
| 4 | -0.21184 | -0.41826 | -0.21614 |
| 5 | -0.36305 | 0.392359 | 0.241791 |
| 6 | 0.183083 | 0.183964 | -0.72608 |
| 7 | -0.18943 | 0.014139 | -0.34516 |
| 8 | -0.2767 | 0.112375 | -0.51494 |
| 9 | -0.23306 | -0.34069 | -0.12797 |
| 10 | -0.54286 | -0.05512 | 0.01827 |
| 11 | 0.143756 | 0.319551 | -0.16369 |
| 12 | -0.45433 | 0.507255 | 0.435882 |
| 13 | 0.079139 | 0.3067 | -0.48478 |
| 14 | -0.15157 | -0.13028 | -0.09997 |
| 15 | -0.44044 | 0.657029 | 0.127414 |

## 4. EXPERIMENTALRESULTS AND OBSERVATIONS

A Neural Network is constructed and simulated from the training dataset.Table 2 shows a matrix of weights originating from layer 2 of the neural network where each row represents a neuron centre and the three columns corresponding to a row represents the three weights from that neuron. Similarly Table 3 shows weights originating from layer 1, the input layer of the neural network where each row represents a neuron centre and the nine columns corresponding to a row represents the nine weights from that neuron. Table 4 shows the bias that is to be added at each neuron centre where each row represents a neuron. The weight and bias matrices were too long to be included in their fullest form here. These tables provide us with a general idea as to how the network behaves.

A large database of images in varied lighting conditions was taken which was used to create the Bayesian Routine and to feed the mentioned input parameters into the three algorithms discussed above. A subset of the Training set is shown below.The input parameters for these images are given in Table 5. Table 6 shows the confusion matrix of all the test images with respect to the neural network, where each cell on the diagonal has the percentage of inputs correctly classified.

Table 4: BIAS MATRIX FOR THE NEURAL NETWORK

| Neuron# | Bias |
|---|---|
| 1 | 2.0715 |
| 2 | -1.9243 |
| 3 | 1.8362 |
| 4 | 1.7118 |
| 5 | 1.2582 |
| 6 | 1.2794 |
| 7 | 1.0699 |
| 8 | -0.9596 |
| 9 | -0.6124 |
| 10 | 0.5830 |

Table 3: WEIGHTS AT THE INPUT LAYER

| Neuron# / Weight# | 1 | 2 | 3 | 4 | 5 | 6 | 7 | 8 | 9 |
|---|---|---|---|---|---|---|---|---|---|
| 1 | -0.1165 | 0.8260 | 0.1047 | -0.0650 | -0.4442 | -0.7919 | 0.8983 | -1.3138 | -0.0930 |
| 2 | 0.7747 | 1.0490 | -0.7296 | -0.1780 | -0.4034 | 0.3964 | 0.1476 | -0.3533 | 1.1496 |
| 3 | -0.6271 | 0.0365 | -0.0245 | 0.7946 | 0.6055 | -0.6684 | -0.8766 | 0.7518 | 0.9167 |
| 4 | -0.3982 | 0.7594 | -0.9261 | -0.7680 | 0.1246 | 0.2520 | 0.1483 | 0.6317 | -1.0628 |
| 5 | 0.6764 | 0.1757 | -1.0113 | -0.8356 | 0.4237 | 0.6529 | 0.7252 | -0.5928 | 0.8112 |
| 6 | -0.1427 | 0.8755 | 1.0259 | -0.8603 | 0.9217 | 0.2421 | 0.3175 | -0.7782 | 0.5010 |
| 7 | -0.3006 | -0.5574 | -0.6255 | 1.2972 | 0.2128 | 0.5115 | -0.7284 | -0.8833 | -0.1306 |
| 8 | -0.6987 | -1.1168 | -0.0092 | 0.6438 | -0.5534 | -0.1377 | -0.5261 | -1.2451 | 0.1030 |
| 9 | 0.2760 | 0.4759 | 0.0258 | -0.8083 | -0.7837 | -0.6376 | -1.2079 | -0.1709 | 0.8297 |
| 10 | -0.7505 | -0.6623 | -0.7850 | 0.4649 | -0.1906 | 0.1311 | 1.0825 | -1.0321 | 0.4333 |

**Table 5:** INPUT PARAMETERS FOR SOME OF THE TRAINING DATA POINTS

| Training Sample# | Mean Hue | Mean Saturation | Mean Value | Mean Y | Mean Cb | Mean Cr | Mean Red | Mean Green | Mean Blue |
|---|---|---|---|---|---|---|---|---|---|
| 1 | 0.162068 | 0.340032 | 0.372549 | 110.34945 | 117.01452 | 139.58895 | 128.38988 | 104.7 | 87.5811 |
| 2 | 0.575144 | 0.244795 | 0.403258 | 73.01415 | 126.95497 | 133.14684 | 74.659666 | 62.58538 | 64.266852 |
| 3 | 0.241863 | 0.578858 | 121.31361 | 132.11358 | 125.73741 | 118.97631 | 122.86689 | 130.94496 | 109.23422 |
| 4 | 0.493929 | 0.241863 | 0.578858 | 121.31361 | 132.11358 | 125.73741 | 118.97631 | 122.86689 | 130.94496 |

The output for some of test cases has been shown in Figure2 and 3. Figure2(a) shows one of the images given as input to all the algorithmsand so is true for Figure3(a). Figure 2(b) shows the output image when the input was given to algorithm 1 i.e. colourspace switch simulating on the Neural Networks and the colour space that it output is RGB for input image in Figure 2(a). Three inset images in Figure 2(b) represent the binary image as given by Bayesian Routine on the colour space chosen by the neural network, the binary image after noise reduction and the corresponding colour image respectively. Figure 2(c) depicts the output when algorithm 2 is applied to the input image. Figure2(d) shows the output when algorithm 3 is applied to input the image. The three inset images in each of the sub-plots are similar to those given for Figure 2(b).

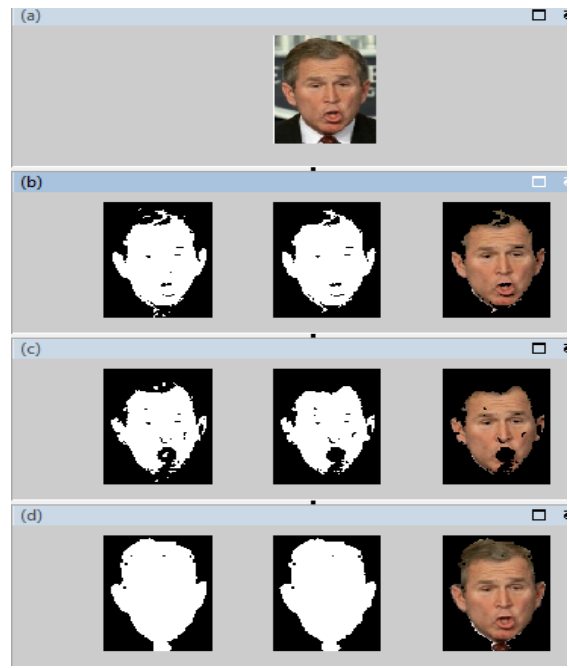

Fig.2.Outputs from after applying all the Algorithms (a)Input Image1(b) Neural Network: RGB. (c)Algorithm2: HSV. (d) Algorithm 3: ∑MaxConnected[i].

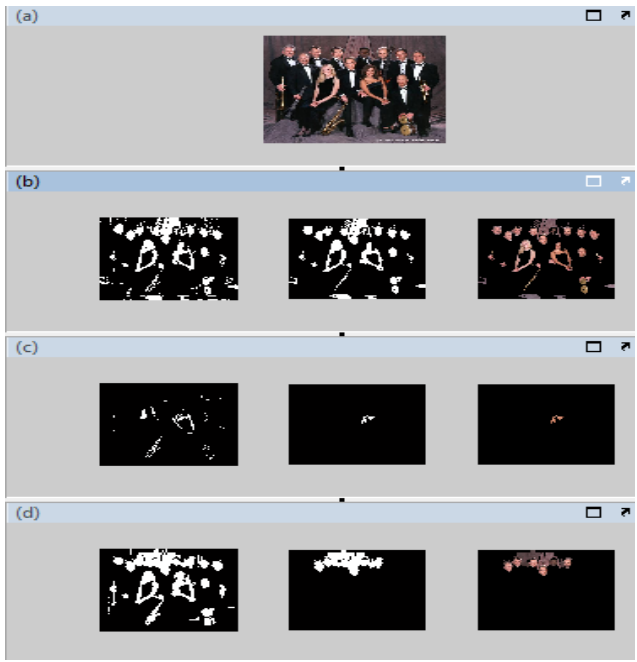

Fig.3Outputs from after applying all the Algorithms (a)Input Image2 (b)Neural Network : YCbCr. (c)Algorithm2:RGB (d) Algorithm 3: ∑MaxConnected[i]

## 5. CODES AND ALGORITHMS

We made test bed formentioned techniques on data such as live feed from a standard webcam installed on a laptop.

### A. *Webcam capture with a delay of 1000000 microseconds*

The following is a pseudo-code for capturing a frame from the standard camera installed on computers. This can be used to train and test the algorithms discussed on an incremental basis (read online training) and to also give this project a logical extension to hand gesture recognition in varied light conditions to control the computer remotely.

```
    Procedure: StreamCaptureAndTrain
1.  start
2.      handle:GetHandle(installed camera)
3.      Stream:Port(handle).communicate(START_STR
    EAM)
4.      Stream.start()
5.      Loop(1000000)
6.      Start
7.              Get_Time()
8.      End
9.      Image:stream.TakeSnapShot()
10.     Stream.stop()
11.     autoSkinSegment(Image)
12. End
```

### B. *BayesianRoutine for RGB Segmentation*

Thispseudo-code loads an image into the workspace and scans its each pixel to determine if the pixel under scanner satisfies Bayesian Probability of belonging to skin class in RGB space. It then goes on to extract the biggest blob of connected segmented skin, using K-Means Clustering, understood to be as the area of interest and then performs filtering on salt and pepper noise present after Bayesianfilter.

```
    Procedure:BayesianRoutine
1.  start
2.  Image:loadImage
3.  for (i,j) in range(sizeof(Image))
4.  do
5.      H:getHue
6.      S:getSaturation
7.      V:getValue
8.      if (
        H ε [0.04, 0.0882]
                S ε [0.11 , 0.68]
                V ε [0.38, 0.112]
            )
9.  startIf
10. SkinPixels(i,j) : [H,S,V]
11.     else
12.             SkinPixels(i,j) : [0,0,0]
13. endIf
14. end
15. Output=SkinPixels.LargestBlob()
16. end
```

## 6. CONCLUSIONS

In this paper, we have proposed a method for skin pixel identification and classificationthat overcomes challenges of lighting changes. We employed three different techniques to adaptively choose skin colour model.MaxConnected technique extracts all the pixels that were found to be in skin class by BayesianRoutine and then extracts the maximum connected area of this image, called the region of interest. The SigmaConnect technique summed over the maximum connected sub images over all the three colour spaces that were provided by the MaxConnected technique. Finally a neural network was created which had 9 input parameters in the form of properties of the image such as mean of all the red values in the image and output the colour space in which the BayesianRoutine would perform the best. Based on our experiments it can be firmly said that the technique of Neural Networks outperformed the Maximum Connected Matrix and the MaxConnected Matrix in case of segmenting skin pixels if the larger picture of variable lighting conditions is taken into account.